\documentclass[letterpaper, 10 pt, conference]{ieeeconf}  

\IEEEoverridecommandlockouts                              

\usepackage{amsmath} 
\usepackage{amssymb}  
\usepackage{graphicx}
\usepackage[subrefformat=parens]{subcaption}

\newif\iflatexe\latexetrue
\title{\LARGE \bf
Casting manipulation of unknown string by robot arm
}

\author{Kenta Tabata$^{1}$, Hiroaki Seki$^{1} $, Tokuo Tsuji$^{1} $, Tatsuhiro Hitamitsu$^{1} $
\thanks{*This work was not supported by any organization}
}

\begin{document}

\maketitle
\thispagestyle{empty}
\pagestyle{empty}

\begin{abstract}

Casting manipulation has been studied to expand the robot’s movable range. In this manipulation, the robot throws and reaches the end effector to a distant target. Usually, a special casting manipulator, which consists of rigid arm links and specific flexible linear objects, is constructed for an effective casting manipulation. However, the special manipulator cannot perform normal manipulations, such as picking and placing, grasping, and operating objects. We propose that the normal robot arm, which can perform normal tasks, picks up an unknown string in the surrounding environment and realizes casting manipulation with it. As the properties of the string are not provided in advance, it is crucial how to reflect it in casting manipulation. This is realized by the motion generation of the robot arm with the simulation of string movement, actual string manipulation by the robot arm, and string parameter estimation from the actual string movement. After repeating these three steps, the simulated string movement approximates the actual to realize casting manipulation with the unknown string. We confirmed the effectiveness of the proposed method through experiments.

\end{abstract}

\section{Introduction}

Dexterous manipulation by robots has been extensively studied for a long time\cite{Rambow}-\cite{yan}. The manipulation of flexible liner objects, such as ropes and cables, is a challenging problem in robotics. They easily deform and the recognition and prediction of movement are challenging. The dynamic manipulation is more affected by the flexible object characteristics. In this study, we focus on casting manipulation as one of the dynamic manipulations with flexible liner objects. In this manipulation, the robot throws and reaches the end effector to a distant target. It has the advantage of expanding the robot workspace. It can also be cast into a narrow space by avoiding obstacles. 

Previous studies have developed special manipulators for casting manipulation. A flexible cable or rope was attached to the tip of the rigid robot arm. Suzuki et al. developed a special casting manipulator consisting of a one-degree-of-freedom (DOF) link and cable. Winding and casting manipulations were analyzed by modeling the cable as a multi-link \cite{Suzuki1}-\cite{Suzuki2}. Arisumi et al. developed a casting equipment on the robot arm end to obtain a sample from the moon crater, where the rover cannot approach. This equipment can launch the penetrator with a cable and reel it up. They proposed a launching method to maintain the posture of the penetrator during casting \cite{Arisumi}. 

Their special manipulator can realize casting manipulation. However, it is not suitable for daily situations. It cannot perform normal manipulations such as picking and placing, grasping, and operation/manipulation of the surrounding items. A robot arm that can perform both basic manipulations and casting manipulations is desired. We propose that the normal robot realizes casting manipulation by picking up a string in the surrounding environment. Therefore, our proposal should address dynamical manipulation of a string with unknown properties.

Some studies have been carried out on the dynamic manipulation of strings, which resembles casting manipulation. Yamakawa et al. formulated an equation of motion to express the string movement and demonstrated that, when one end of the string is grasped and moved at a high and constant speed, the string motion follows the trajectory of the robot arm. They used this to achieve dynamic string operation and cloth folding operations \cite{Yamakawa2016}-\cite{Yamakawa2010}. Sawada et al. developed a mass–spring model, which includes the bending properties that vary depending on the string elongation and realize the casting manipulation by a 1-DOF link \cite{Sawada}.  Several results of dynamic manipulation of strings have been reported. However, these studies assumed that the string properties are obtained by special tests in advance. Multiple types of strings with different properties were not considered.

In our study, after the normal robot arm grasped a string, the robot manipulated the string to reach the tip to the target position. To realize casting manipulation, we propose a method of parameter estimation of the string from the actual string movement in casting manipulation trials. This method does not require many identification tests in advance. 

	\begin{figure}[h]
		\centering
		\includegraphics[width=0.85\hsize]{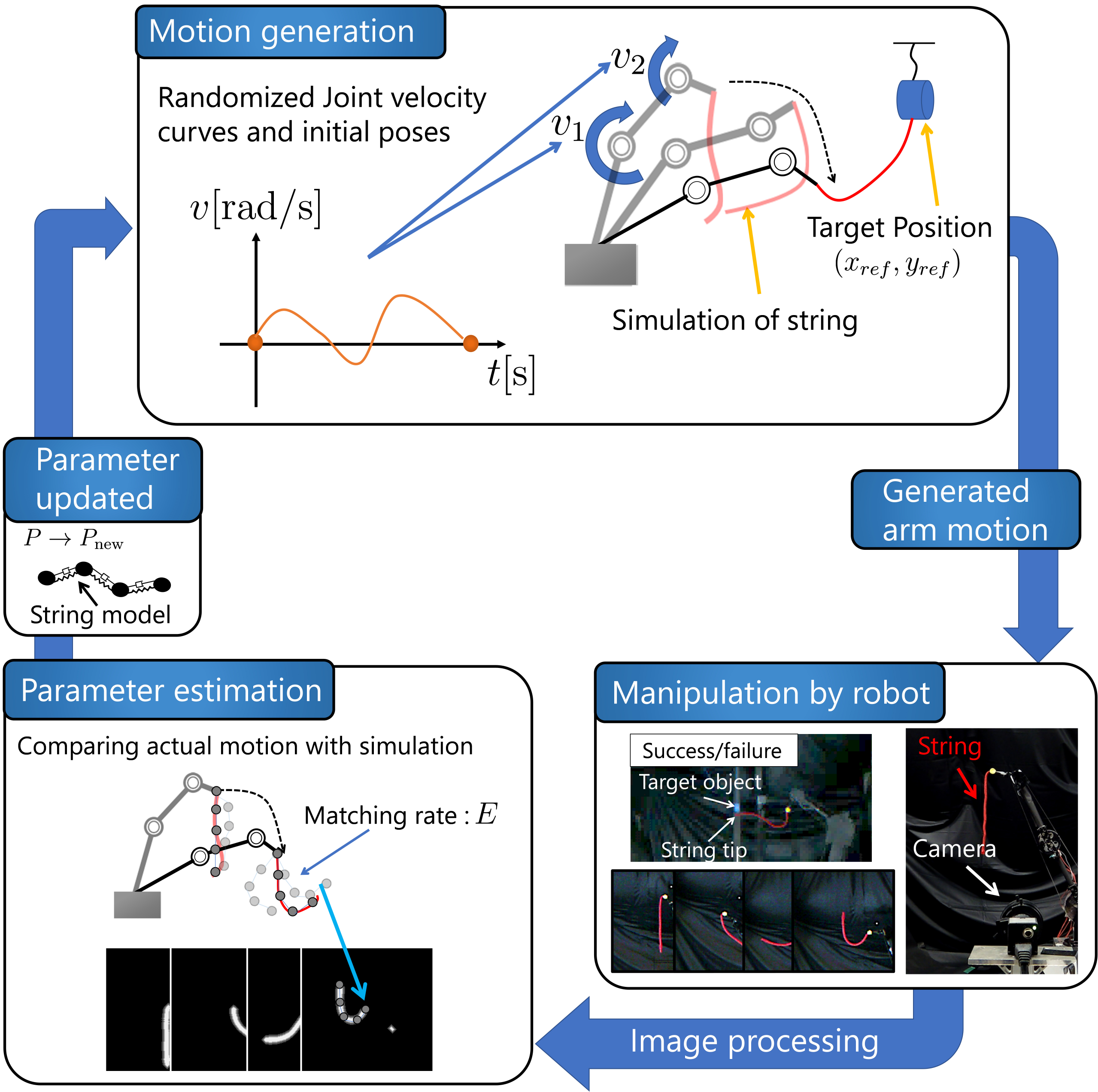}
		\caption{Proposal concept of casting manipulation for unknown string}
		\label{fig:concept}
	\end{figure}

\section{Casting manipulation of an unknown string}

\subsection{Proposal for realizing casting manipulation}

Casting manipulation is realized with a 3-DOF robot arm in a two-dimensional (2D) space. The robot arm grasped the unknown string and manipulated it to reach the tip to the target position. We propose a method for this casting manipulation, as shown in {\bf{Fig.\ref{fig:concept}}}.

First, the target position is provided. The initial parameters of the string model were set randomly. In motion generation, the robot arm movement is based on the joint angular velocity. This movement of the robot arm was generated randomly and the initial arm position was randomly determined within the movable range. The movement of the string was simulated from the grasp point movements of the robot arm using a mass–spring–damper model. When the simulated string tip reached the target position, the motion generation was completed.  

Second, the generated motion was performed by an actual robot arm. The manipulation was filmed using a camera. Image processing was used to extract the motion of the string alone, which was then saved.

In the first manipulation, the string model parameters and actual string properties do not match. Therefore, we estimated the string parameters. By providing string parameters randomly, we simulated the string motion based on the actual arm movement. The matching rates of the simulated and actual string movements were analyzed. The parameter combination with the highest matching rate was retained. From the second manipulation onward, motion generation and actual manipulation were performed using the estimated string parameters. 

By repeating this procedure, the actual and simulated string movements gradually approach each other. Manipulation was then generated to reflect the string property and realize casting manipulation. If the casting manipulation is not achieved after repeating this procedure several times, the manipulation is regarded as a failure.
	\begin{figure}[h]
		\centering
		\includegraphics[width=0.78\hsize]{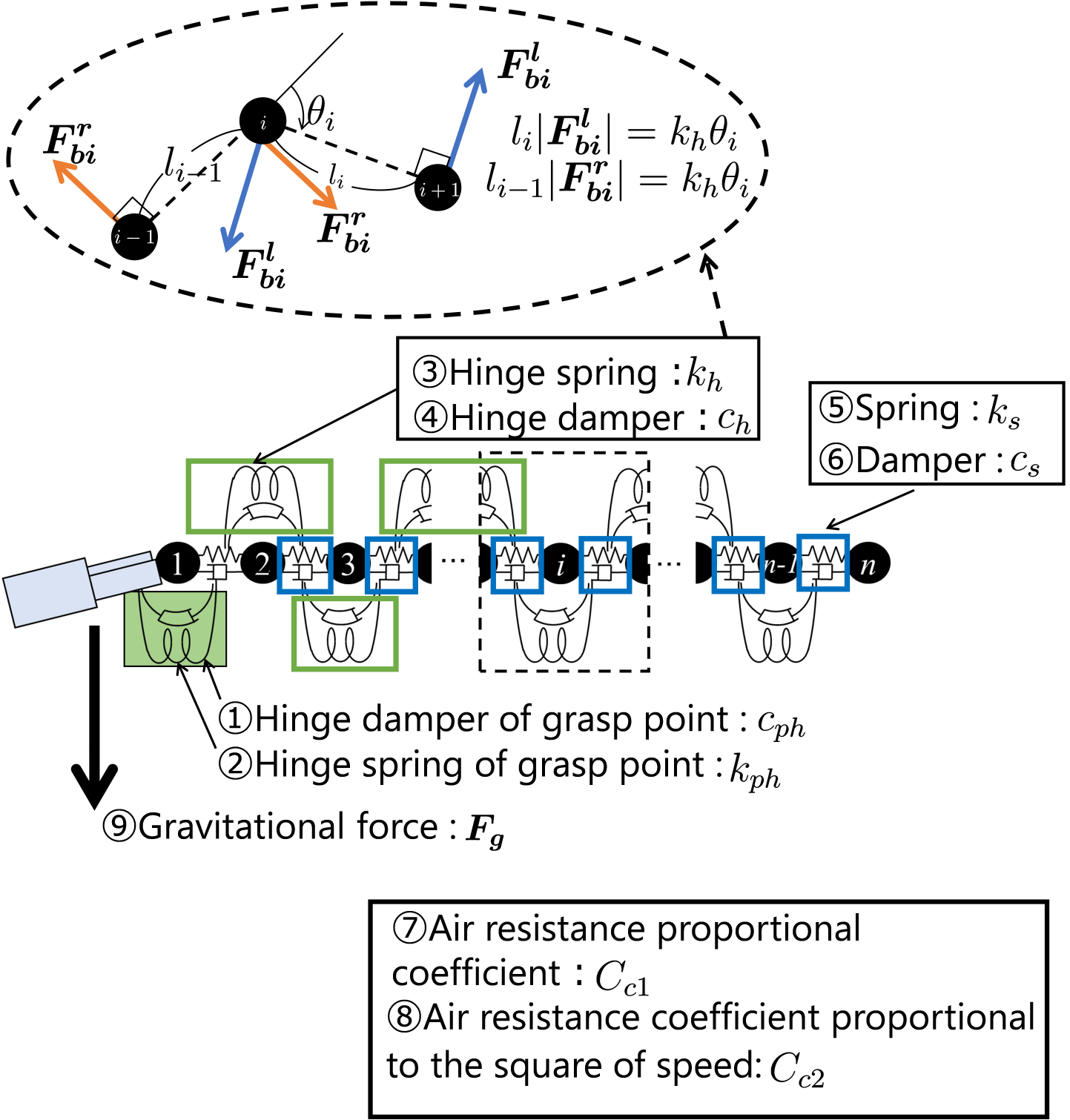}
		\caption{String model}
		\label{fig:string_model}
	\end{figure}

\subsection{String model}

The proposed method requires repeating the string movement simulation for motion generation and parameter estimation. A string model is required for this simulation. The mass–spring–damper model was selected because of its low computational load. Our method does not aim to express the various string movements completely, but only a specific string movement for casting manipulation.

We assume that the string is homogeneous. Twisting is not considered because if the string movement and observation plane are limited to 2D, the effect of twisting is also contained in 2D. To represent the properties of elongation and bending, the string model is composed of mass points, springs, dampers, hinge springs, and hinge dampers, as shown in {\bf{Fig.2.}} The mass point numbers were set to $i=1,..,n$, starting from the grasping point. The equation of motion for mass point $i$(Position vector $\vec{\ddot{{\bf{r}}}}_{i}$) is expressed as follows: 

		\begin{align}
			m \vec{\ddot{{\bf{r}}}}_{i}&=\vec{F}_{si}-\vec{F}_{si-1}+\vec{F}_{di}-\vec{F}_{di-1}+\vec{F}^{r}_{bi-1}-\vec{F}^{r}_{bi}-\vec{F}^{l}_{bi} \notag \\ 
				& \qquad+\vec{F}^{l}_{bi+1} +\vec{F}^{r}_{hi-1}-\vec{F}^{r}_{hi} -\vec{F}^{l}_{hi}+\vec{F}^{l}_{hi+1}\notag \\ 
				& \qquad \qquad +\vec{F}_{p}+\vec{F}_{g}+\vec{F}_{ci} \\
			\vec{F_{p}} &=
				\begin{cases}
					\vec{F}_{ph}+\vec{F}_{phc}, & \text{if } i=1\\
					0, & \text{otherwise}
				\end{cases}
			\label{eq:string}
		\end{align}

\begin{center}
\begin{table}[b]
	\caption{String model parameters}
	\label{tb:model_variable}
	\scalebox{0.75}
	{
	\begin{tabular}{|c|c|c|}
		\hline 
		Force & Coefficient & Explanation of parameter \\ \hline
		$F_{g}$ &$k_{s}$ &\begin{tabular}{c}Elastic force \\between the mass points\end{tabular}\\ \hline
		$F_{d}$ &$c_{s}$&\begin{tabular}{c}Damping force\\ between the mass points\end{tabular}\\ \hline
		$F_{b}$ &$k_{h}$&\begin{tabular}{c}Force caused by torsional spring moment\\ between the three mass points\end{tabular}\\ \hline
		$F_{h}$ &$c_{h}$&\begin{tabular}{c}Force caused by torsional damper moment\\ between the three mass points\end{tabular}\\ \hline
		$F_{c}$ &$C_{c1}, C_{c2}$(Squared term)&\begin{tabular}{c}Air resistance at the mass point \end{tabular}\\ \hline
		$F_{g}$ & - &Gravitational force\\ \hline
		$F_{ph}$ & $k_{ph}$&\begin{tabular}{c}Torsional spring moment between\\ the robot hand and grasped mass point\end{tabular}\\ \hline
		$F_{phc}$ &$c_{ph}$&\begin{tabular}{c}Force caused by torsional damper moment between\\ the robot hand and grasped mass point\end{tabular}\\ \hline
	\end{tabular}
	}
\end{table}
\end{center}

The forces and their coefficients are shown in {\bf{Table\ref{tb:model_variable}}}. Both sides of the Eq. (1) are divided by the mass $m$ of the mass points and unit mass conversion (i.e., designation of a value in $k_{s}/m$ to the spring constant) is performed for each parameter. Thus, we do not need to consider the mass; there will be eight string parameters ($k_{h}$, $c_{h}$, $k_{s}$, $c_{s}$, $C_{c1}$, $C_{c2}$, $k_{ph}$, and $c_{ph}$). 

When the string is manipulated by a robot arm, time-series data about the orientation and position of the robot finger (the first mass point coordinates $\vec{r}_{1}$) are provided. The positon vector $\vec{{{\bf{r}}}}_{i}$ for each string mass point is obtained by a numerical calculation (Euler's method) of the equation of motion, which is the simulation of the string movement.
\newpage
\section{Robot arm motion generation for casting manipulation}

	\begin{figure}[h]
		\centering
		\includegraphics[width=\hsize]{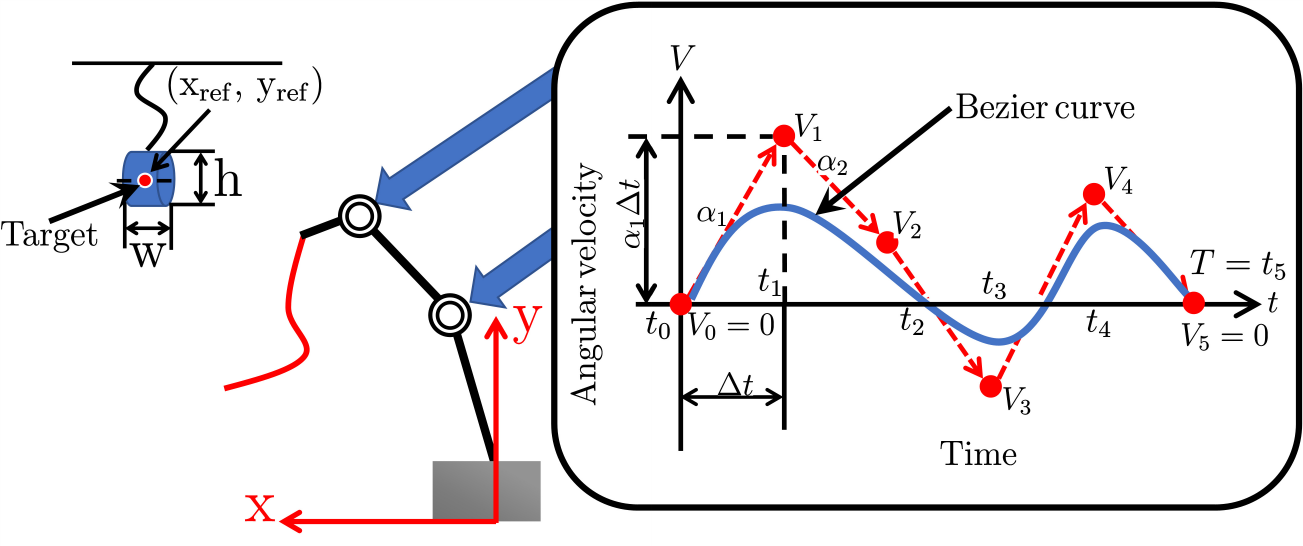}
		\caption{Motion generation by joint velocity curve}
		\label{fig:hit_set}
	\end{figure}

The initial angles for each joint of the robot arm were randomly selected from the movable range and specified as the initial positions. Subsequently, a joint velocity curve was generated using a Bezier curve, as illustrated in {\bf{Fig.\ref{fig:hit_set}}}. Time $T$ is determined randomly in a certain range (e.g., 0.2 s$\sim$1.5 s), and the time from 0 to T is divided into five equal parts ($t_{0}\sim t_{5}$). The acceleration $\alpha_{k}$ at time $t_{k}$ is randomly determined from the range of limit accelerations of the robot arm. We consider that the robot moves at a uniform speed in each time interval $\Delta t$. The joint velocity for the control points is determined by $V_{0}=V_{5}=0$. 
	\begin{align}
		V_{n}=\alpha_{n} (t_{n}-t_{n-1}) \qquad (n=1, \cdot \cdot \cdot,4)
	\end{align}
Using the control points $V_{0} \sim V_{5}$, a Bezier curve is generated and used as the joint velocity curve. This was performed for all joints. We confirmed that the last generated movement of the arm did not exceed the limits of the robot’s movable range or speed. When there are no problems, a simulation from the robot arm motion to the string motion is performed. 

The target object was placed at ($x_{ref}$, $y_{ref}$)  with an allowable error of $(w, h)$. When the string tip enters the target object area in the simulation, the motion generation is terminated. If the string tip does not reach the target object during the robot arm movement, new initial arm position and joint velocity curve are generated. This is repeated until the casting manipulation is achieved. 

For the second and subsequent motion generations, the initial position generated during the first time was unchanged. The previously generated velocity curve was used for the joint velocity curve with a slight change. The motion finish time $T$ is randomly changed within 1/2 of the range of the first and is based on the previous value. The joint velocity for each of the control points $V_{1} \sim V_{4}$ is randomly changed based on the previous value and in approximately a fraction of 1 of the range of the previous value. The achievements in subsequent simulations were judged in the same manner as that of the first.

\section{String parameter estimation}

The actual string movement in the casting manipulation was captured by the camera and used for parameter estimation. The image of the string is extracted based on the grasping point of the robot arm. The value for each string model parameter was selected randomly. The motion simulation of the string was performed using the robot arm motion. The matching rate $E$ is calculated by comparing the point positions in the string model obtained from the simulation and image series of the actual string motion. This was repeated while changing the parameters. After a fixed number of repetitions, the eight parameters with the highest matching rates were output as the estimated parameters. 

\subsection{Random parameter selection}
When randomly selecting each parameter in the string model, its value was determined using the exponential form. This allows the parameter range to vary widely. For a fast parameter convergence, we narrow the parameter estimation range in a stepwise manner using the following equations:. For manipulation times M, the number of parameter changes shall be m and a certain parameter shall be $P_{a}$.

	\begin{align}
		P_{a}&=P_{min}\left( \dfrac{P_{max}}{P_{min}} \right)^{\chi_{m}}, \, 0\leq \chi_{m} \leq 1 \\
		\chi_{m} &= \chi_{\mathrm best} + \dfrac{\chi_{0}}{M} \beta^{m} \cdot RAND(-1,1)
	\end{align}
The maximum and minimum parameter values $P_{max}$ and $P_{min}$, respectively, were determined in advance. The initial value $\chi_{0}$ when determining $\chi_{m}$ was chosen. RAND(-1,1) expresses random numbers -1$\sim$1. $\beta$ is a value slightly below 1 and is used to narrow the search range every time the parameters are updated.  $\chi_{best}$ is the final estimated parameter value (exponent) in the previous manipulation. 

\subsection{Calculating the matching rate}
	\begin{figure}[b]
		\centering
		\includegraphics[width=\hsize]{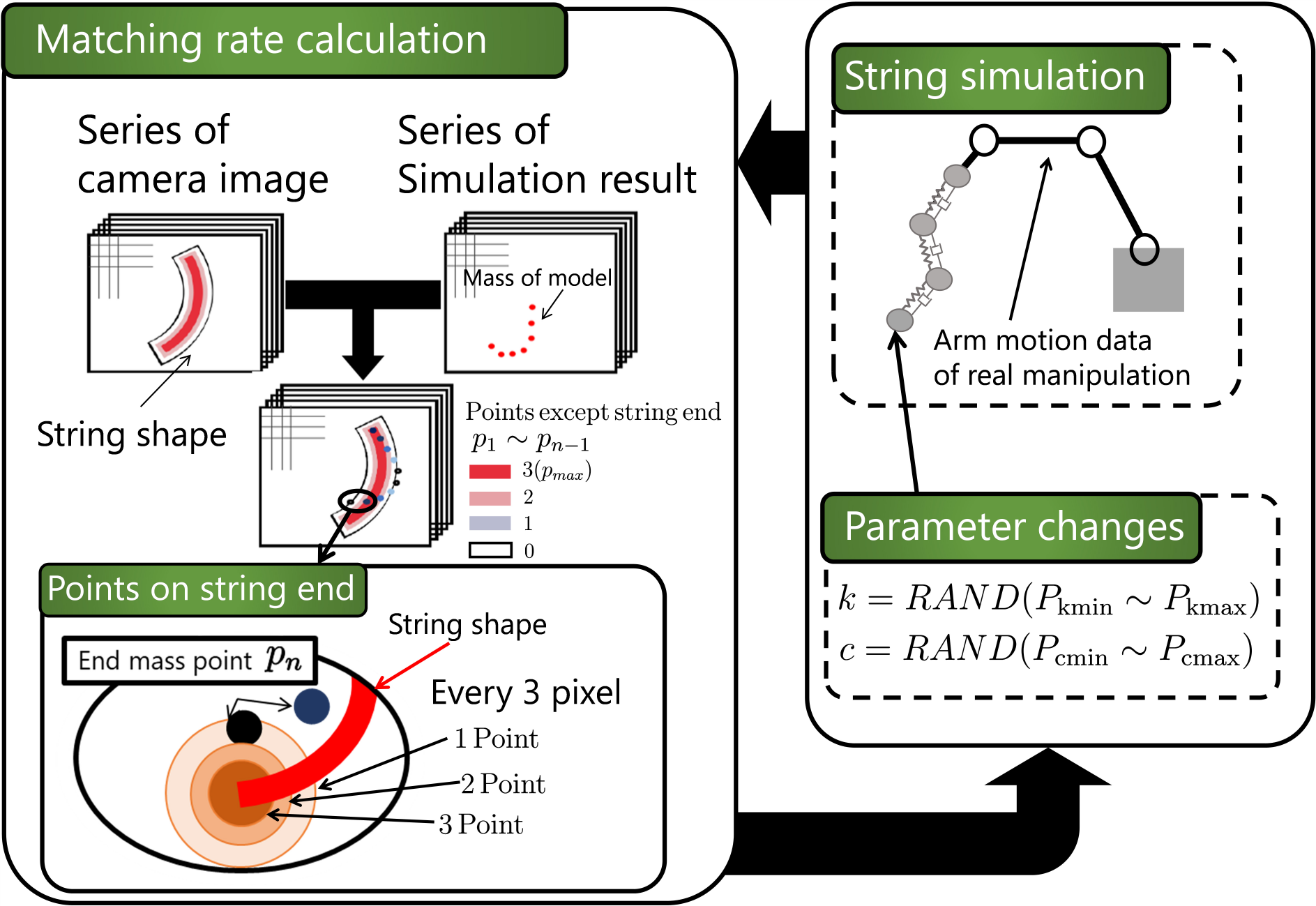}
		\caption{Parameter estimation with matching rate E}
		\label{fig:estimation_loop}
	\end{figure}

The matching rate $E$ is obtained by comparing the actual string movement and string movement simulated by the selected parameter sets, as shown in {\bf{Fig\ref{fig:estimation_loop}}}. The binarized images of the actual string movement are dilated multiple times and weighted scores are assigned in accordance with the application times of dilation. Thus, a closer area to the center corresponds to a higher score ($pmax,...2, 1, 0$). The mass point positions in the string model correspond to the score $p_{i}$ based on the expanded area to which they belong.  

The score of the string tip is specially calculated because the tip position of the string is important in casting manipulation. The above scoring method does not guarantee that the simulated string tip approaches the actual tip position. Therefore, the actual string tip position is obtained by a depth-first search from the grasp point. The score of the simulated string tip (mass point) was assigned based on the distance (every three pixels) from the actual string tip.

At this point, the scores are weighted depending on the mass point number ($i = 1,..., n$), because the mass points near the grasping area are likely to move slightly; they do not contribute to the parameter estimation. The movement increases near the end of the string. Hence, the weighting  $w_{i}$ increases toward the end of the string. After weighting each image $E_{f}$, the sum for all images ($f$=1,..., fmax) becomes the matching rate $E$ using the following equation, where the weighting increment is $\Delta w$.

	\begin{align}
		E&=\dfrac{1}{f_{max}}\cdot{\displaystyle \sum_{f=1}^{fmax}}E_{\mathrm f}, \quad
		E_{\mathrm f}=\dfrac{{\displaystyle \sum^{n}_{i=1}}p_{i}\cdot w_{i}}{p_{\mathrm max}\cdot{\displaystyle \sum^{n}_{i=1}} w_{i}}\\
		w_{i}&=1+(i-1)\Delta w
	\end{align}

	\begin{figure}[h]
		\centering
		\includegraphics[width=0.8\hsize]{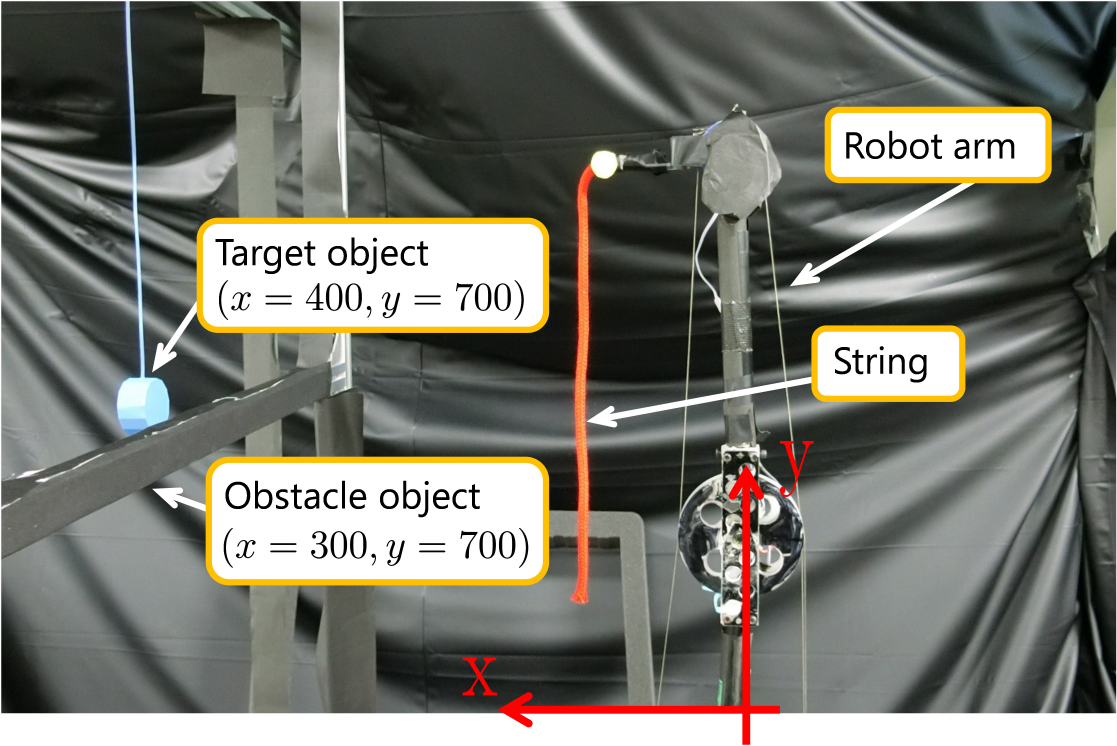}
		\label{fig:avoid_set}
		\caption{Experimental setup with obstacle object}
	\end{figure}
	
	\begin{figure}[h]
		\centering
		\includegraphics[width=\hsize]{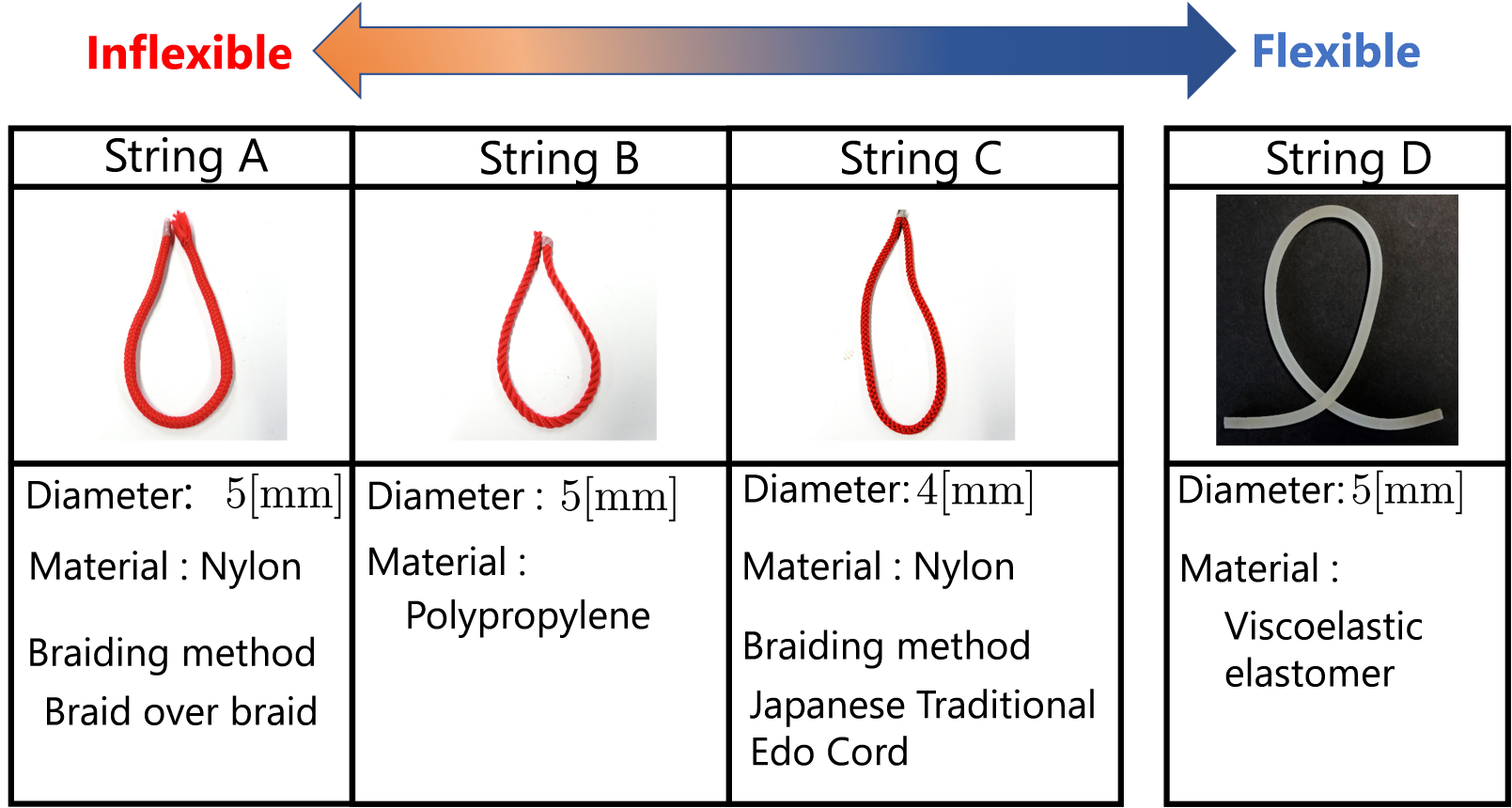}
		\caption{Strings used for experiment}
		\label{fig:string}
	\end{figure}

	\begin{table}[h]
			\caption{Range for parameter estimation}
			\label{tb:parastring}
					\scalebox{0.8}{
				\begin{tabular}{|c|c|c|} \hline
					Parameter coefficient&Minimum&Maximum\\ \hline
					Spring $k_{s}$[N/m/kg] &$9.0 \times 10^{3}$&$9.0 \times 10^{5}$\\ \hline
					Damping $c_{s}$[Ns/m/kg] &$0.13 $&$1.3\times 10^{3}$\\ \hline
					Hinge spring $k_{h}$[Nm/rad/kg]&$8 .0\times 10^{-3}$ & $4.0\times 10^{2}$ \\ \hline
					Hinge damping $c_{h}$[Nms/rad/kg]&$3.0 \times 10^{-7}$ & $0.67$\\ \hline
					Air resistance (proportional) $C_{c1}$ [Nms/rad/kg]&$1.0 \times 10^{-4}$ & 10 \\ \hline
					Air resistance (squared proportional) $C_{c2}$ [Nms/rad/kg]&$1.0 \times 10^{-4}$ & 10 \\ \hline 
					Hinge spring at grasp point $k_{ph}$ [Nm/rad/kg] &$1.0 \times 10^{-3}$& 5.0 \\ \hline
					Hinge damping at grasp point $c_{ph}$ [Nms/rad/kg] & $1.1 \times 10^{-6}$ & $0.37$ \\ \hline
				\end{tabular}
				}
	\end{table}
\section{Experiments}

We investigated whether casting manipulation could be achieved using the proposed method. We created a 3-DOF wire-driven robot arm for this manipulation, as shown in {\bf{Fig.5}}. The total length of the robot arm was 585 mm, while the maximum composite speed of the hand was 21.8 m/s. The arm was moved by providing velocity commands every 5 ms. 

The length of the manipulated string was 300 mm. Four types of strings were prepared, as shown in {\bf{Fig.\ref{fig:string}}}: String B was harder than String A, String C was softer than String A, and String D was softest. The string model was given ten mass points ($n$ = 10). The range for selecting the robot movement time is $T=0.2 \sim 1.5$s. {\bf{Table\ref{tb:parastring}}} lists the ranges of all parameters used in the parameter estimation. The initial values used for the first motion generation are the minimum values presented in this table. Furthermore, the convergence factor is $\beta=0.995$,the initial search range is $\chi_{w0}=0.6$, and the weighting increment used for the matching rate is $\Delta w=0.25$. The actual manipulation image series used for parameter estimation are the image series added every 0.2 s before and after movement time $T$. In this study, we defined the allowable error as (w, h) = (0.02, 0.04) [m].

	\begin{figure}[h]
		\begin{minipage}{0.49\hsize}
		\centering
		{
		\includegraphics[width=\hsize]{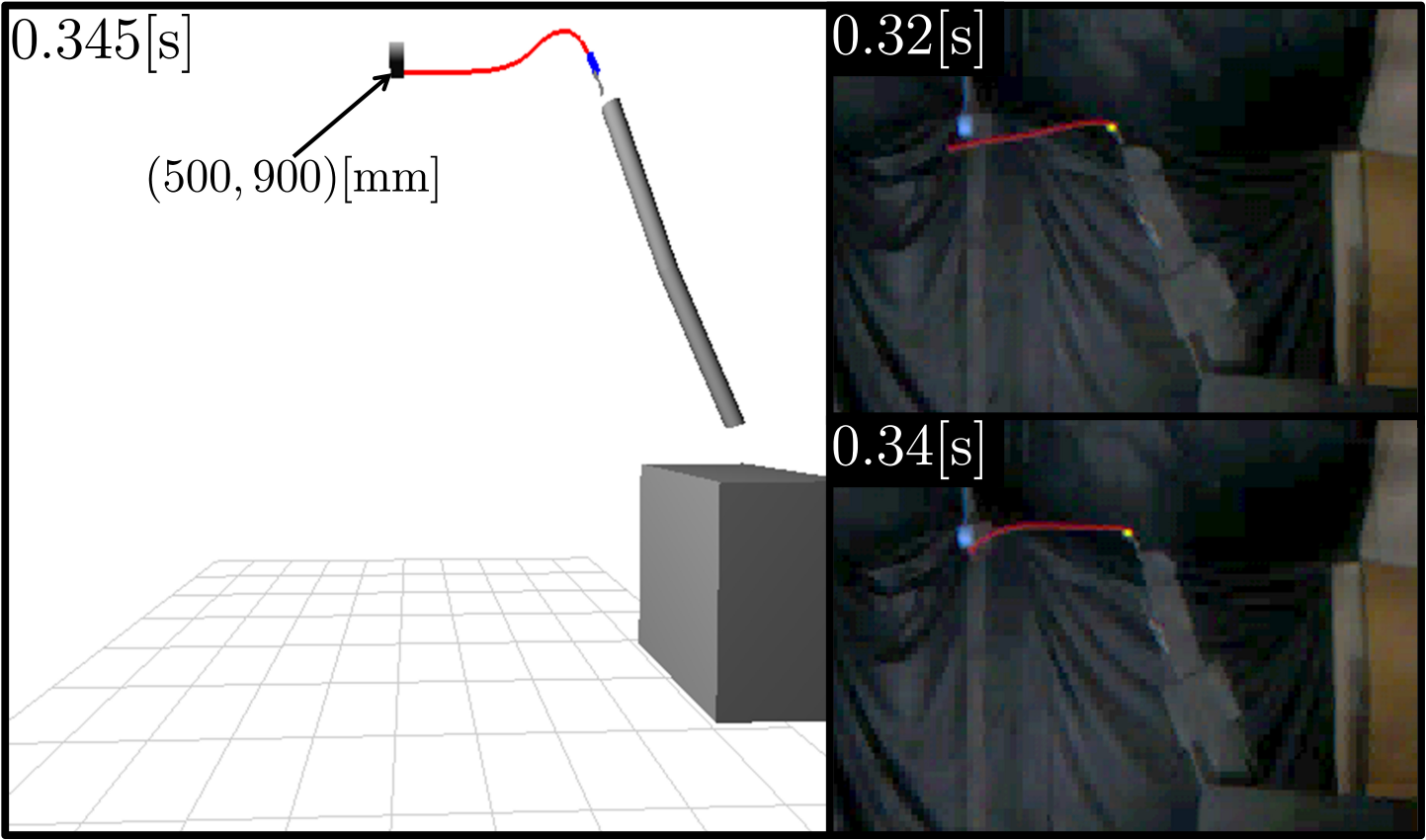}
		\subcaption{1st manipulation}
		\label{fig:manipu1-1}
		}
		\end{minipage}
		\begin{minipage}{0.49\hsize}
		{
		\centering
		\includegraphics[width=\hsize]{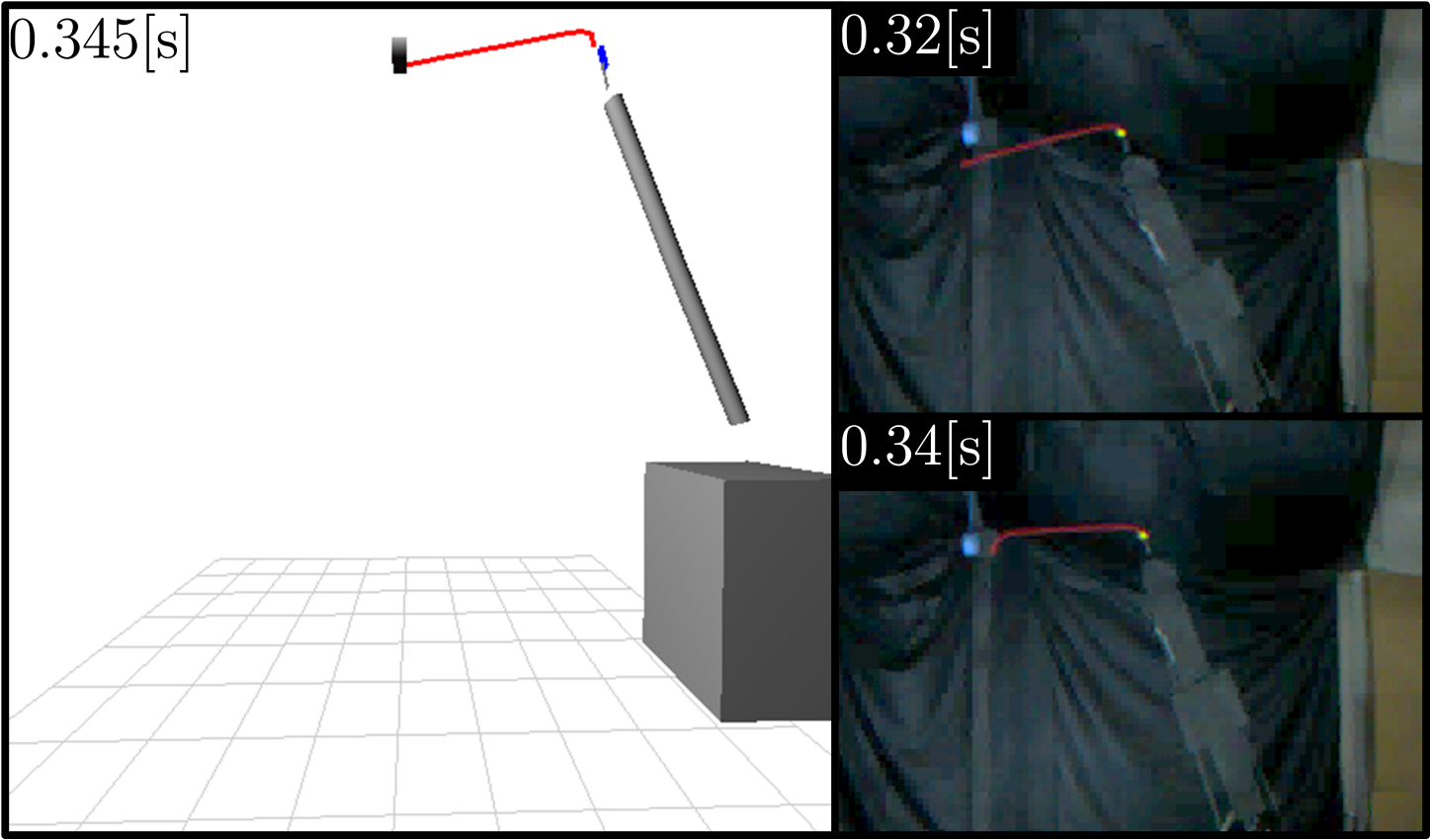}
		\subcaption{2nd manipulation}
		\label{fig:manipu1-2}
		}
		\end{minipage}
		\caption{Casting manipulation with stringA to (500, 900)}
		\label{fig:strAA}
	\end{figure}

	\begin{figure}[h]
		\begin{minipage}{0.49\hsize}
		\centering
		{
		\includegraphics[width=\hsize]{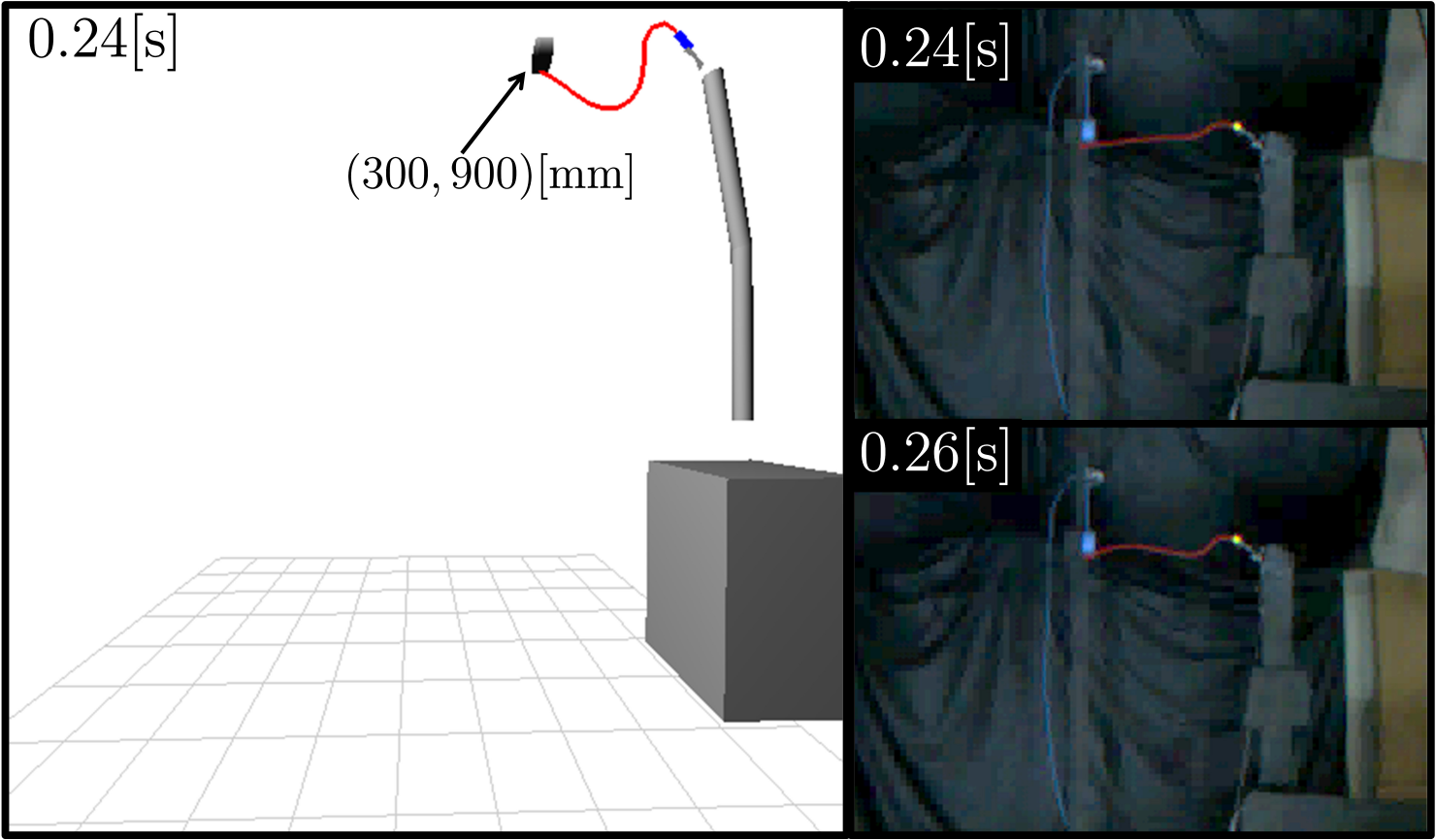}
		\subcaption{1st manipulation}
		\label{fig:manipu3-1}
		}
		\end{minipage}
		\begin{minipage}{0.49\hsize}
		{
		\centering
		\includegraphics[width=\hsize]{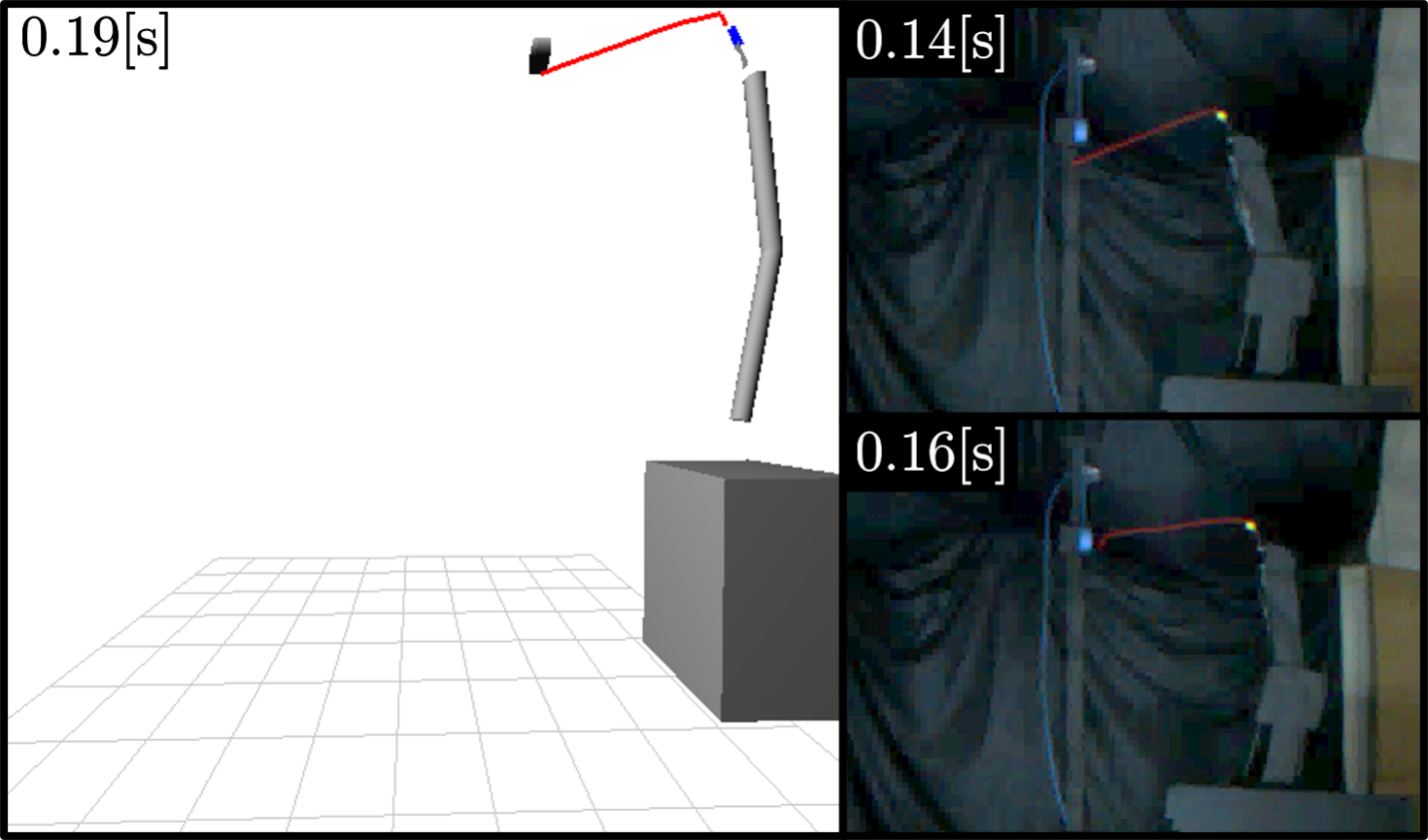}
		\subcaption{2nd manipulation}
		\label{fig:manipu3-3}
		}
		\end{minipage}
		\caption{Casting manipulation with stringC to (300, 900)}
		\label{fig:strCC}
	\end{figure}
\newpage
	\begin{figure}[b]
		\begin{minipage}{0.49\hsize}
		\centering
		{
		\includegraphics[width=\hsize]{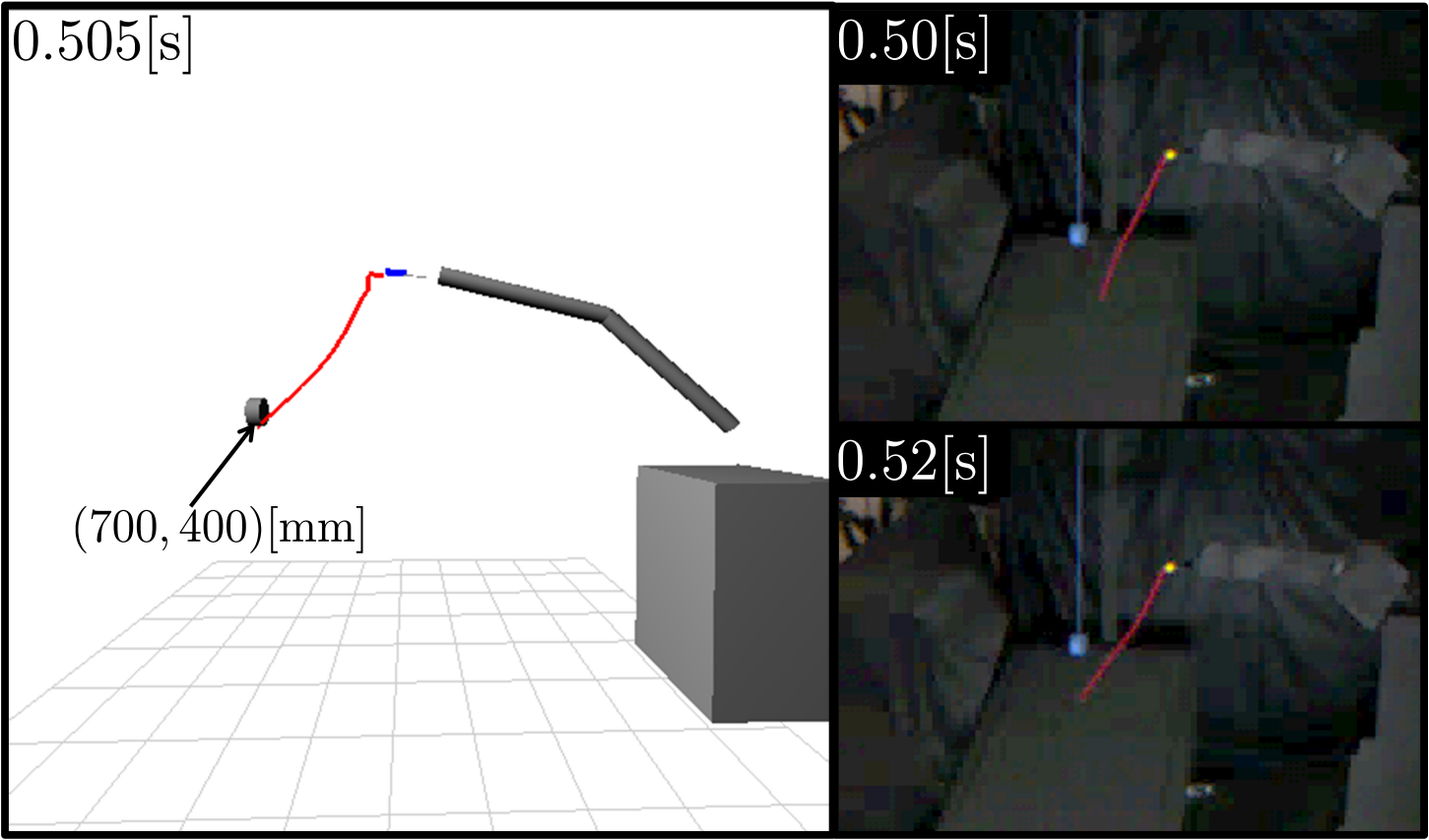}
		\subcaption{1st manipulation}
		\label{fig:manipu2-2}
		}
		\end{minipage}
		\begin{minipage}{0.49\hsize}
		{
		\centering
		\includegraphics[width=\hsize]{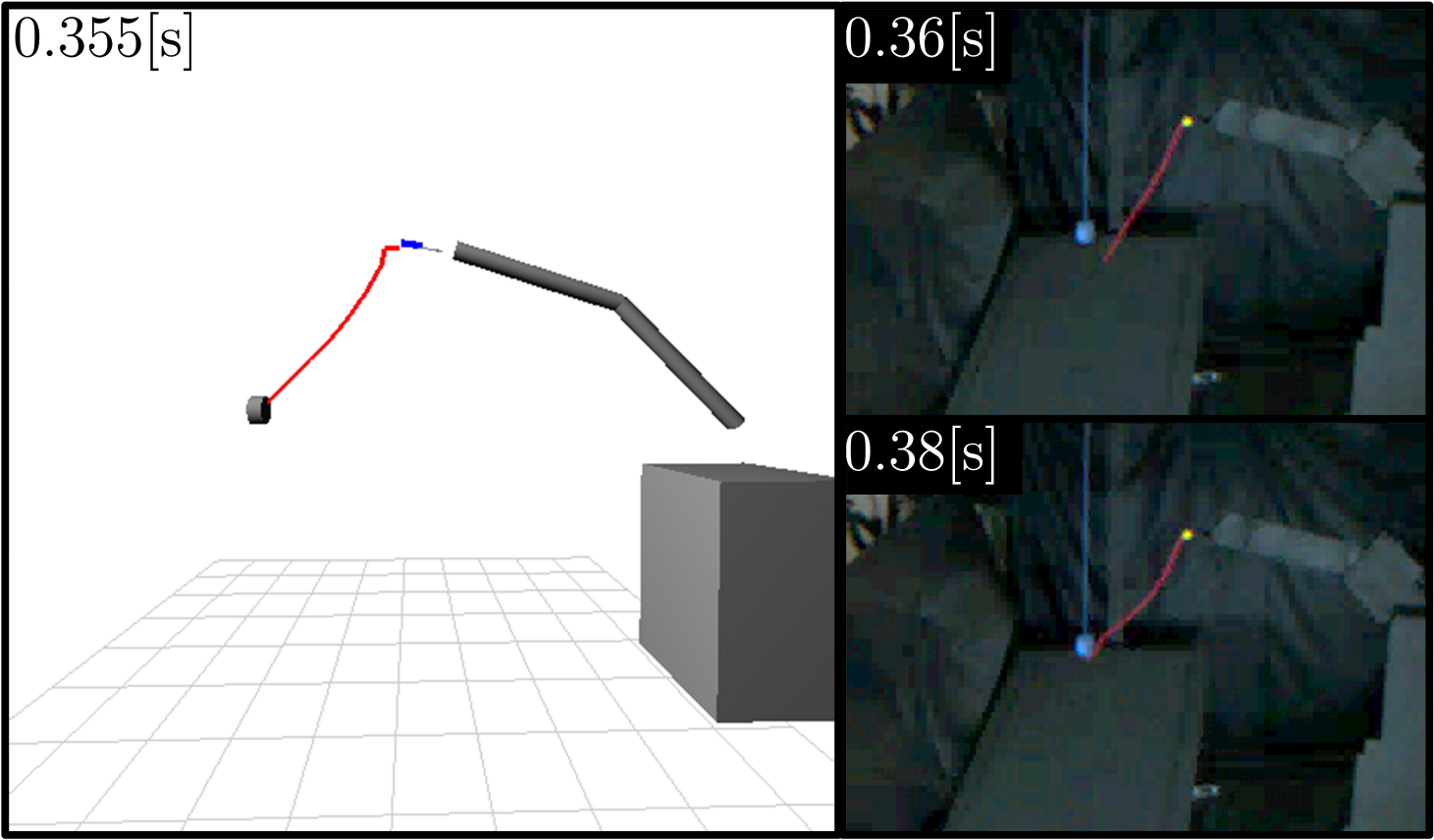}
		\subcaption{2nd manipulation}
		\label{fig:manipu2-3}
		}
		\end{minipage}
		\caption{Casting manipulation with stringB to (400, 700)}
		\label{fig:strBB}
	\end{figure}

\begin{figure}[b]
		\centering
		\includegraphics[width=\hsize]{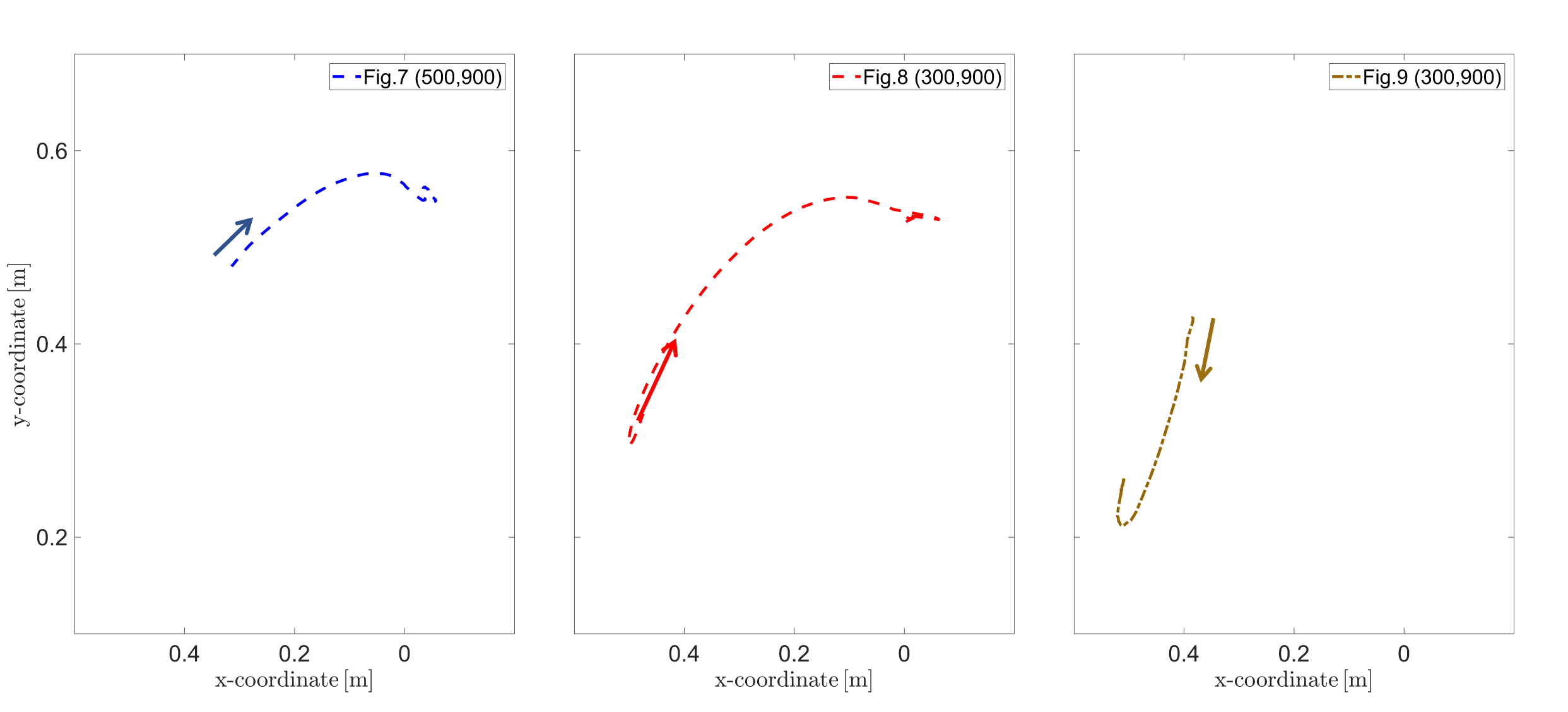}
		\caption{Arm tip trajectory}
		\label{fig:comp_tra}
	\end{figure}

\subsection{Casting manipuilation to different target position}

We investigated whether casting manipulation with an unknown string could be achieved in a variety of target positions. We set three target positions. {\bf{Fig.\ref{fig:strAA}}, \ref{fig:strCC}, \ref{fig:strBB}} show the casting manipulations with strings A, B, and C, respectively. The panel (a) in each figure shows the generated motion with the initial parameter and actual first manipulation, while the panel (b) shows the results of the second manipulation. 

As shown in {\bf{Fig.\ref{fig:strAA}}} and {\bf{Fig.\ref{fig:strCC}}},the first manipulation could achieve the casting manipulation, but the string shape was different from the simulation at the achieved moment. The actual shape became the same as that of the simulated string after the string parameter estimation.

As shown in {\bf{Fig.\ref{fig:strBB}}}, casting manipulation could not be achieved in the first trial. After the parameter estimation, the string tip reached the target object. 

The arm tip trajectories in each manipulation are illustrated in {\bf{Fig.\ref{fig:comp_tra}}}. The initial pose and trajectory of the robot arm differed in their manipulation. In the case of the target positions of (300, 900) and (400, 700), the arm tip moves as swinging up. In contrast, the arm tips move as swinging down when the target position is (900, 500). This implies that the generated motion changes depend on the target position.

\subsection{Casting manipulation with strings with different properties}

\begin{figure}[h]
		\centering
		\includegraphics[width=0.8\hsize]{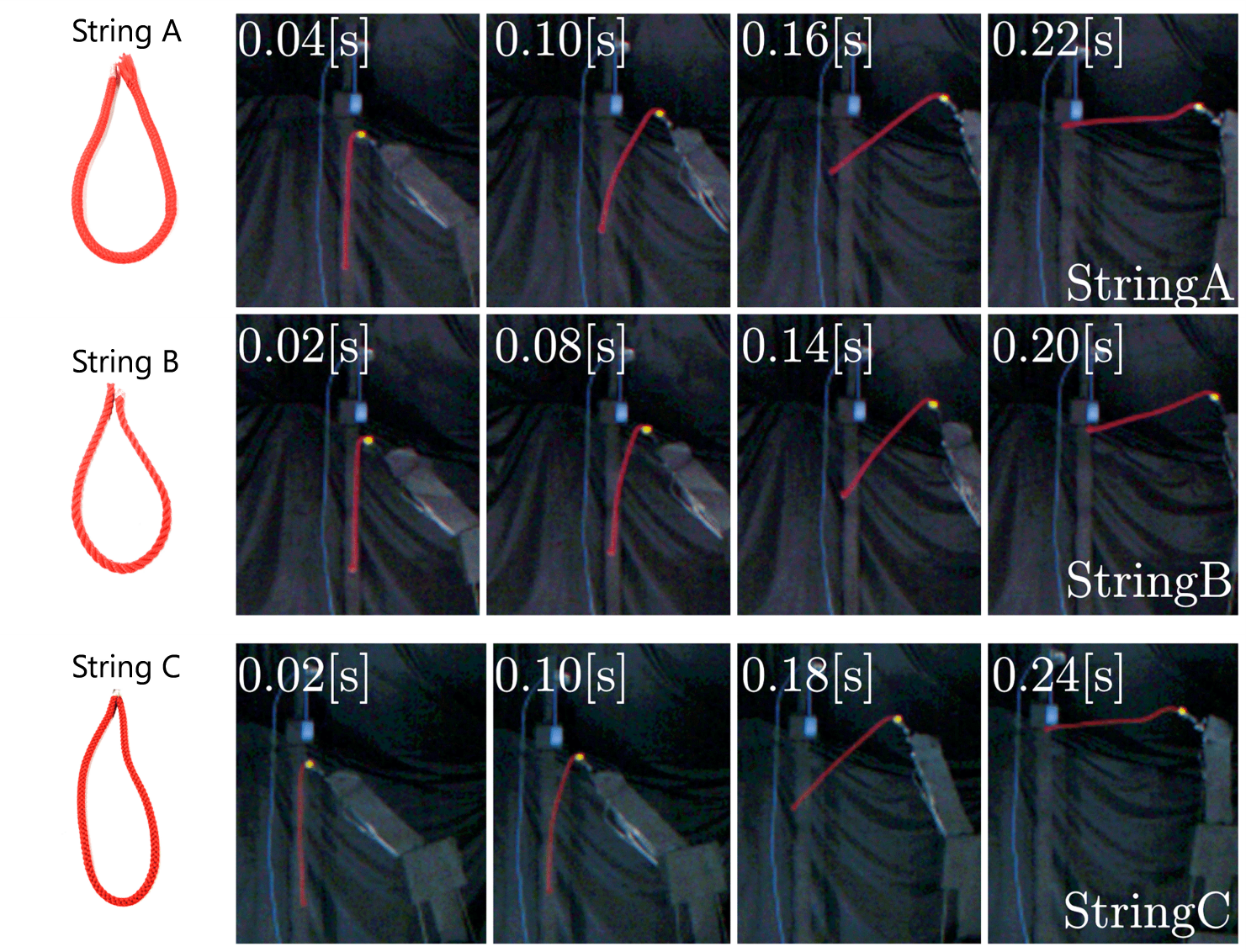}
		\caption{Achieved casting manipulation with different strings}
		\label{fig:comp_movement}
	\end{figure}

Casting manipulation was performed using strings A, B, and C. We set the target position to ($\rm{x_{ref}}$, $\rm{y_{ref}}$)=(300, 900) mm. The same generated motion was used for the first manipulation with all strings. We confirmed that all manipulations were achieved once by repeating the proposed method, as shown in {\bf{Fig\ref{fig:comp_movement}}}. {\bf{Table\ref{tb:estimated}}} lists the representative estimated parameters for each string. $k_{h}$ and $c_{h}$ were estimated the minimum and maximum values, respectively, in the estimation range. As the string did not bend significantly in these manipulations, $k_{h}$ and $c_{h}$ had a low sensitivity to the string movement. In contrast to these results, $k_{ph}$ and $c_{ph}$ were estimated appropriately because the part of the string near the grasp point bent more. 

Additionally, a more flexible string, D, was tested. The number of mass points was changed from 10 to 25 to express the more flexible shape. The target position was set to (${\rm{x_{ref},y_{ref}}}$)=(500,500) mm. As a result, casting manipulation was performed ({\bf{Fig. 12}}). Casting manipulation with a complex characteristic string requires a sufficient expression of the string movement in the simulation.

	\begin{table}[h]
			\caption{Estimated string parameters}
			\label{tb:estimated}
					\scalebox{0.8}{
				\begin{tabular}{|c|c|c|c|c|}\cline{3-5}
					\multicolumn{1}{c}{}&\multicolumn{1}{c|}{}&\multicolumn{3}{c|}{Estimated value}\\ \cline{1-5}
					Parameter&Initial value&string A&string B&string C \\ \hline
					$k_{h}$[Nm/rad/kg]&$8\times10^{-3}$&$8\times10^{-3}$&$8\times10^{-3}$&$8\times10^{-3}$\\ \hline
					$c_{h}$[Nms/rad/kg]&$1.92\times10^{-7}$&0.19&0.10&0.19\\ \hline
					$k_{ph}$[Nm/rad/kg]&$1.0\times10^{-3}$&44.7&0.80&0.57\\ \hline
					$c_{ph}$[Nms/rad/kg]&$6.8\times10^{-7}$&$1.4\times10^{-4}$&0.15&$4.5\times 10^{-3}$\\ \hline
				\end{tabular}
				}
	\end{table}

	\begin{figure}[h]
		\centering
		\includegraphics[width=\hsize]{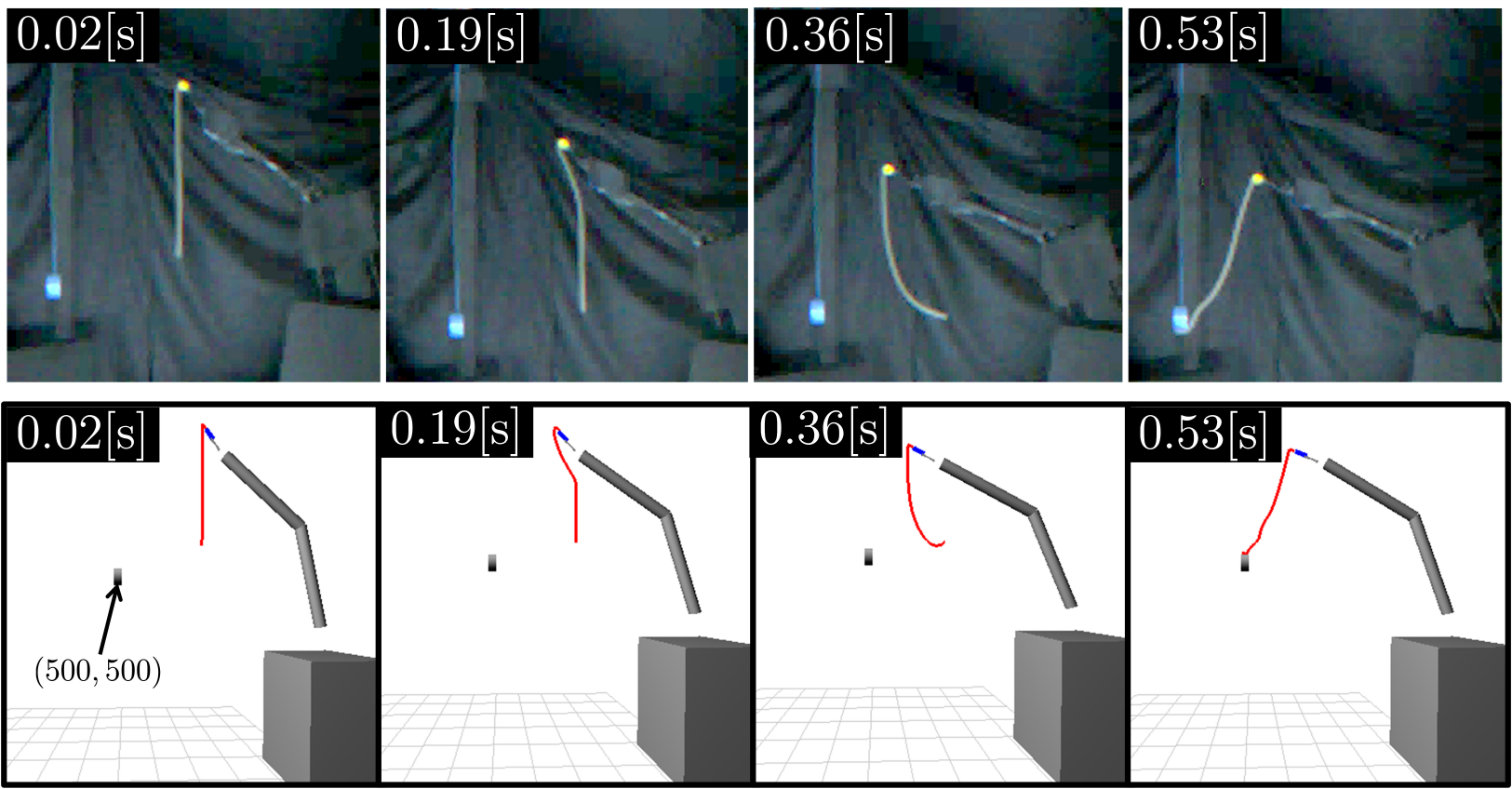}
		\caption{Casting manipulation with stringD}
		\label{fig:manipu4-1}
	\end{figure}

\subsection{Casting manipulation avoiding an obstacle}

Finally, we analyze the casting manipulation in an environment with obstacles, as shown in {\bf{Fig.5}}. A wall is an obstacle object between the robot arm and target object.

\begin{figure}[h]
		\begin{minipage}{\hsize}
		\centering
		{
		\includegraphics[width=\hsize]{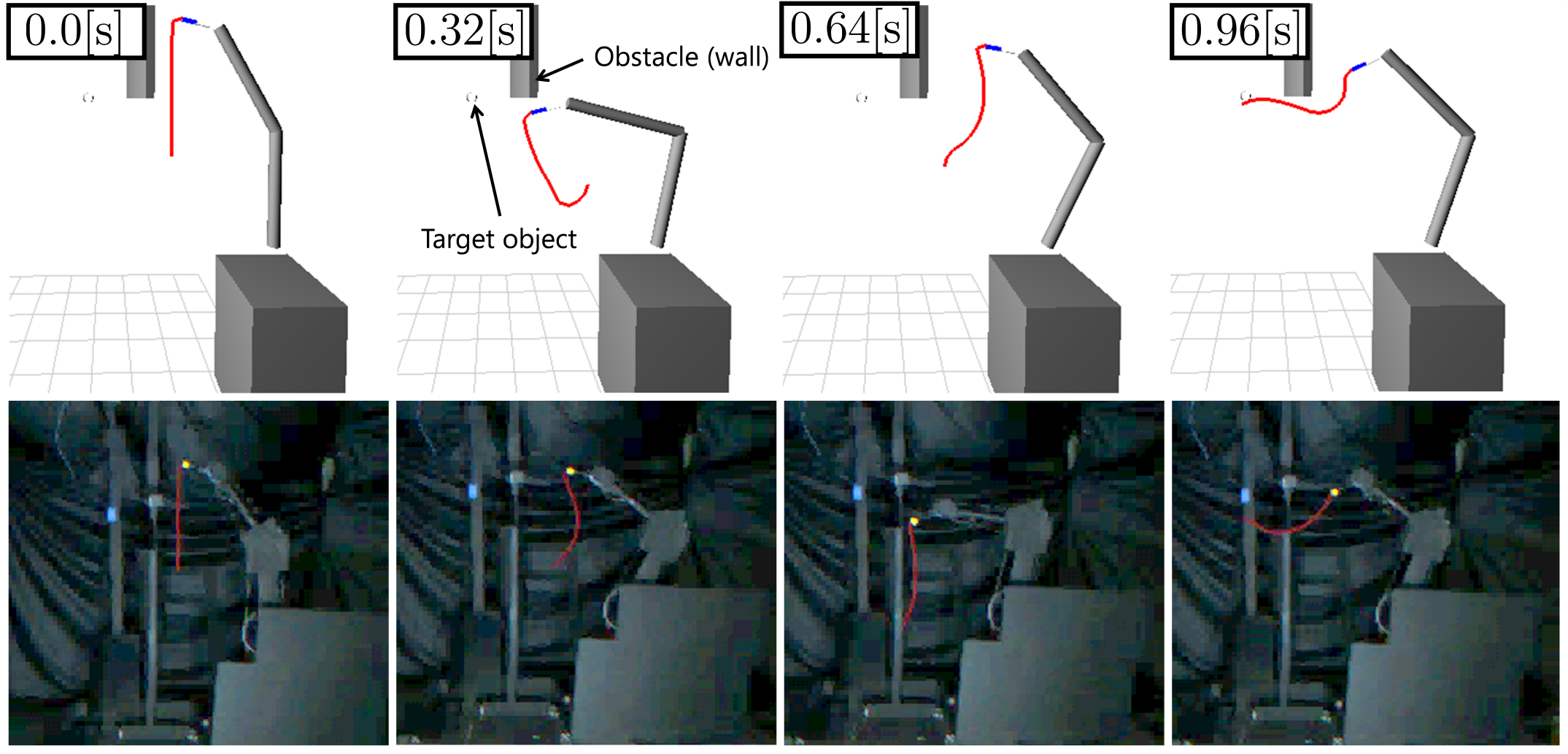}
		\subcaption{1st manipulation}
		\label{fig:avoid1st}
		}
		\end{minipage}
		\begin{minipage}{\hsize}
		{
		\centering
		\includegraphics[width=\hsize]{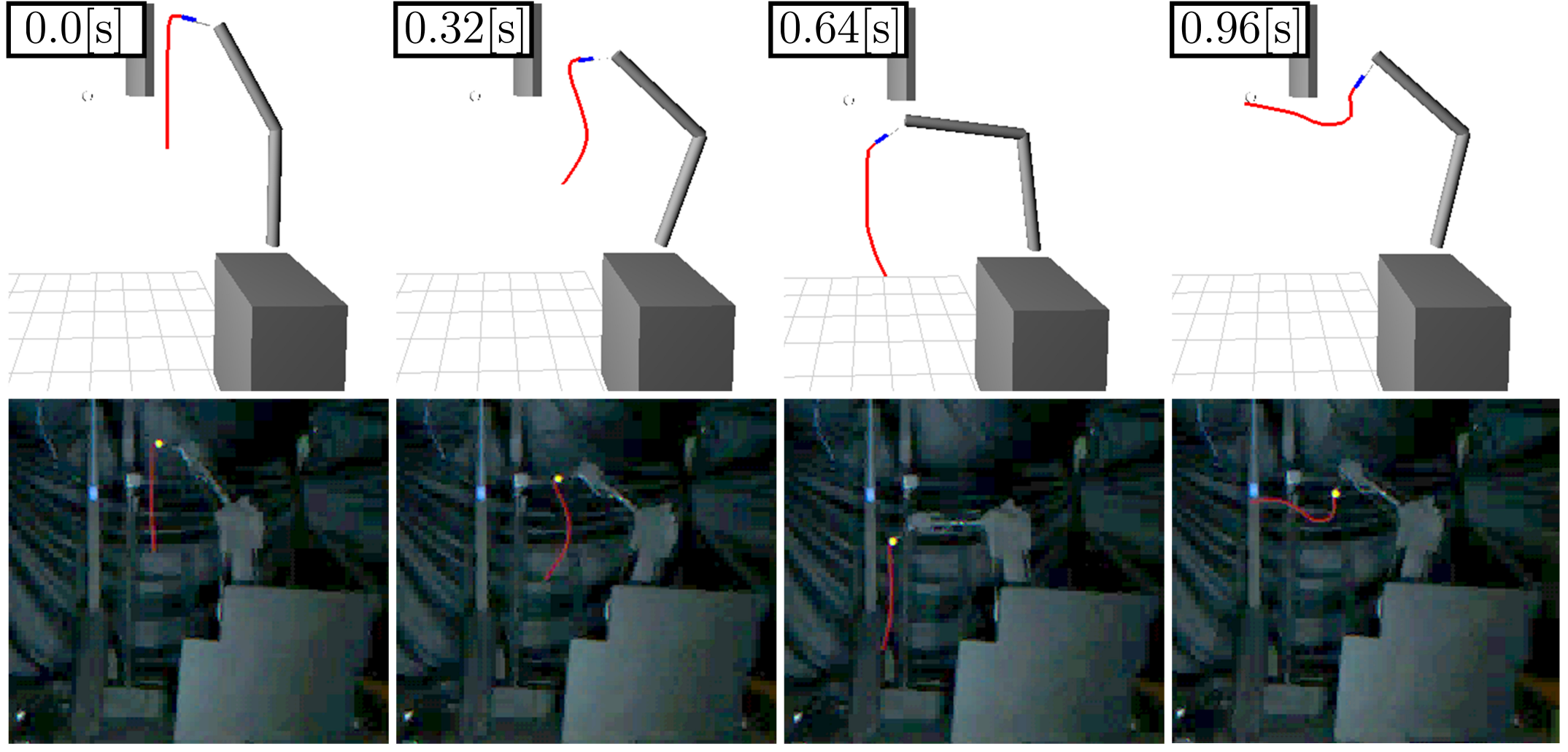}
		\subcaption{5th manipulation}
		\label{fig:avoid5th}
		}
		\end{minipage}
		\caption{Casting manipulation with obstacle object}
		\label{fig:avoid_manipu}

		\centering
		\includegraphics[width=0.8\hsize]{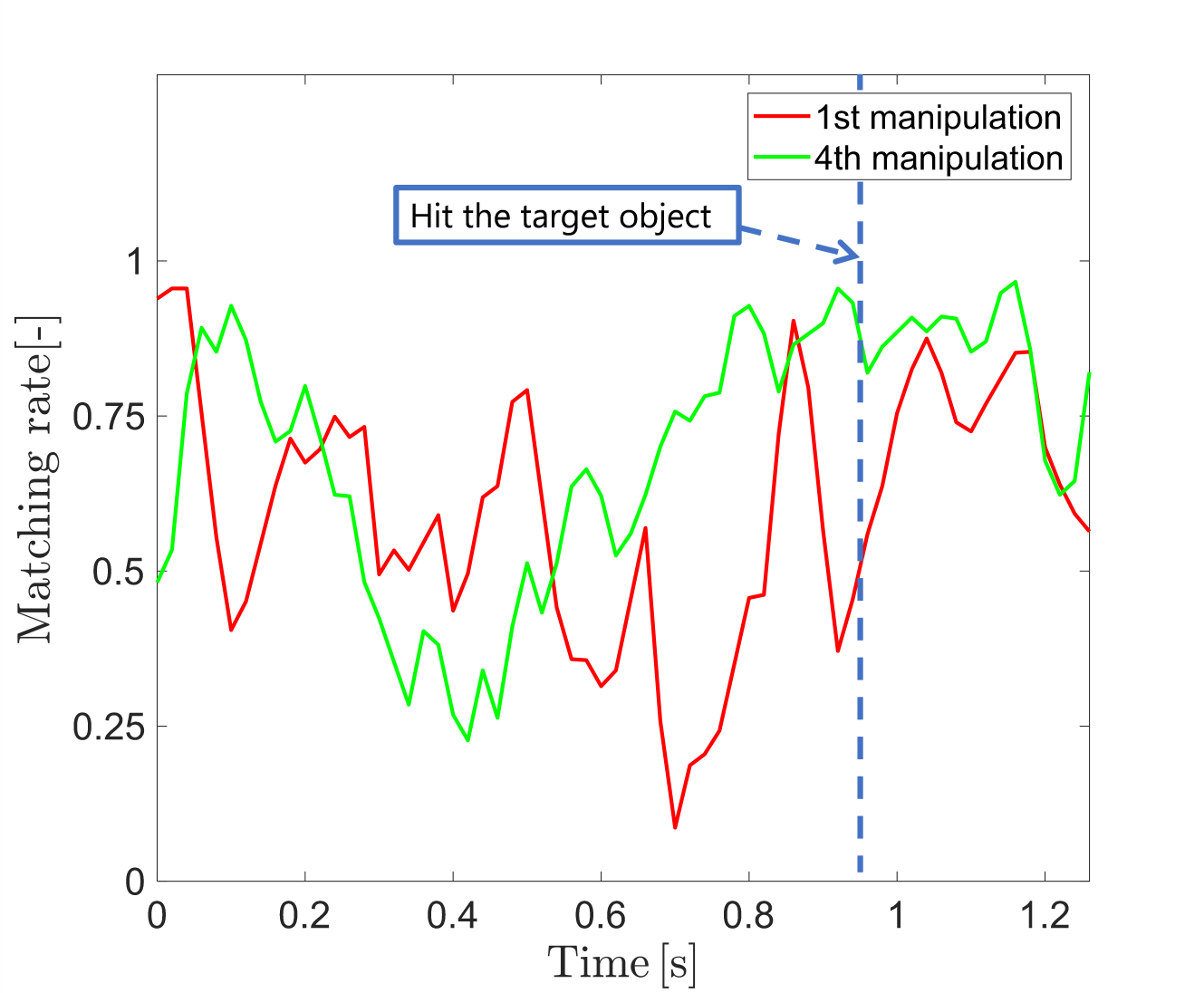}
		\caption{Comparing Matching rate $E$ between 1st and 4th manipulation}
		\label{fig:matchingrate}
	\end{figure}

The casting manipulation was performed after five repetitions({\bf{Fig.\ref{fig:avoid_manipu}}}). The string was manipulated to be bent to avoid the obstacle. Large bending has nonlinear properties; however, the used model is linear. In such cases, our method requires several estimation times to obtain a better casting manipulation. The result of the comparison of the matching rates $E$ of the first and fourth manipulations is shown in {\bf{Fig.\ref{fig:matchingrate}}}. The matching rate $E$ of the fourth manipulation is higher than that of the first manipulation around the achieved moment of casting manipulation. We confirmed that a better estimated parameter was obtained by repeating the motion generation, actual manipulation, and parameter estimation.

\section{Conclusion}
In this paper, we propose a method for casting manipulation of a robot arm with an unknown string. After repeating the three steps with a mass–spring–damper string model, motion generation of the robot arm, actual casting manipulation, and string parameter estimation, the simulated string motion approaches the actual motion and the generated robot motion succeeds the casting manipulation. We tested three different target positions and four types of strings and realized casting manipulation in all cases. The arm motion was generated depending on the string properties and relative position between the robot arm tip and target. This method does not require an identification test of the string for the model parameters in advance. It estimates the string parameters by casting manipulation trials. The manipulation in an environment with an obstacle was also analyzed and achieved after several repetitions of the proposed method. In future studies, we will attempt to realize casting manipulation under more complex situations, such as narrow space, many obstacles, and long strings.



\begin{thebibliography}{99}

\bibitem{Rambow} M. Rambow, T. Schauß, M. Buss and S. Hirche, "Autonomous manipulation of deformable objects based on teleoperated demonstrations," IEEE/RSJ International Conference on Intelligent Robots and Systems, 2012.

\bibitem{Eyoshida} E. Yoshida, K. Ayusawa, Ramirez-Alpizar I G, Harada K, Duriez C, and Kheddar A, Simulation-based optimal motion planning for deformable object, IEEE International Workshop on Advanced Robotics and its Social Impacts , 2015, pp. 1-6.

\bibitem{Alverz} N. Alvarez, K. Yamazaki, T. Matsubara T, An approach to realistic physical simulation of digitally captured deformable linear objects. In: IEEE International Conference on Simulation, Modeling, and Programming for Autonomous Robots , pp. 135-140, 2016.

\bibitem{Lai} Y. Lai, J. Poon, G. Paul, H. Han and T. Matsubara, "Probabilistic Pose Estimation of Deformable Linear Objects," IEEE 14th International Conference on Automation Science and Engineering (CASE), pp. 471-476, 2018.

\bibitem{yan} M. Yan, Y. Zhu, N. Jin and J. Bohg, "Self-Supervised Learning of State Estimation for Manipulating Deformable Linear Objects," in IEEE Robotics and Automation Letters, vol. 5, no. 2, pp. 2372-2379, 2020.

\bibitem{Suzuki1} T. Suzuki, Y. Ebihara, T. Suzuki, Y. Ando and M. Mizukawa, ``Casting and Winding Manipulation with Hyper-Flexible Manipulator," IEEE/RSJ International Conference on Intelligent Robots and Systems, Beijing, pp. 1674-1679, 2006.

\bibitem{Suzuki2} T. Suzuki and Y. Ebihara, "Casting Control for Hyper-Flexible Manipulation," Proceedings IEEE International Conference on Robotics and Automation, pp. 1369-1374, 2007.

\bibitem{Arisumi} H. Arisumi, M. Otsuki and S. Nishida, "Launching penetrator by casting manipulator system,"  IEEE/RSJ International Conference on Intelligent Robots and Systems, Vilamoura, pp. 5052-5058, 2012.

\bibitem{Yamakawa2016} Y. Yamakawa, A. Namiki and M. Ishikawa, ``Simplified deformation model and shape generation of a rhythmic gymnastics ribbon using a high-speed multi-jointed manipulator," Mechanical Engineering Journal, 2016.

\bibitem{Yamakawa2010-2} Y. Yamakawa, A. Namiki and M. Ishikawa, ``Motion planning for dynamic knotting of a flexible rope with a high-speed robot arm," IEEE/RSJ International Conference on Intelligent Robots and Systems, pp. 49-54, 2010.

\bibitem{Yamakawa2010} Y. Yamakawa, A. Namiki and M. Ishikawa, ``Dynamic manipulation of a cloth by high-speed robot system using highspeed visual feedback," Proceedings of the 18th International Federation of Automatic Control, pp.8076--8081, 2011.

\bibitem{Sawada} Y. Sawada and T. Watanabe, ``Casting behavior of the string structure considering axial elogation," Journal of the Japan Society of Mechanical Engineers (in Japanese), 2019.


\end{thebibliography}
\end{document}